\pdfoutput=1

\documentclass[11pt]{article}

\usepackage[preprint]{acl}

\usepackage{times}
\usepackage{latexsym}

\usepackage[T1]{fontenc}

\usepackage[utf8]{inputenc}

\usepackage{microtype}

\usepackage{inconsolata}

\usepackage{graphicx}

\usepackage{amsmath} %
\usepackage{multicol} %
\usepackage{graphicx}  %
\usepackage{array}     %
\usepackage{lipsum}    %

\usepackage{float} %

\newcommand{\camerareadytext}[1]{\xspace}
\newcommand{\ignore}[1]{}
\newcommand{\theword}{fidelity\xspace}
\newcommand{\theworda}{intermediary\xspace}

\usepackage{xspace}
\newcommand{\num}{13\xspace}

\title{The Noisy Path from Source to Citation:\\Measuring How Scholars Engage with Past Research}

\author{Hong Chen \\
  University of Michigan \\
    Ann Arbor, MI, USA \\
  \texttt{hongcc@umich.edu} \\\And
  Misha Teplitskiy \\
  University of Michigan \\
  Ann Arbor, MI, USA \\
  \texttt{tepl@umich.edu} \\\And
  David Jurgens \\
  University of Michigan \\
    Ann Arbor, MI, USA \\
  \texttt{jurgens@umich.edu} \\}

\begin{document}
\maketitle
\begin{abstract}
Academic citations are widely used for evaluating research and tracing knowledge flows. Such uses typically rely on raw citation counts and neglect variability in citation types. In particular, citations can vary in their \textit{\theword} as original knowledge from cited studies may be paraphrased, summarized, or reinterpreted, possibly wrongly, leading to variation in how much information changes from cited to citing paper.
In this study, we introduce a computational pipeline to quantify citation fidelity at scale. Using full texts of papers, the pipeline identifies citations in citing papers and the corresponding claims in cited papers, and applies supervised models to measure \theword at the sentence level. Analyzing a large-scale multi-disciplinary dataset of approximately \num million citation sentence pairs, we find that citation \theword is higher when authors cite papers that are 1) more recent and intellectually close, 2) more accessible, and 3) the first author has a lower H-index and the author team is medium-sized.
Using a quasi-experiment, we establish the ``telephone effect''---when citing papers have low \theword to the original claim, future papers that cite the citing paper and the original have lower \theword to the original. 
Our work reveals systematic differences in citation \theword, underscoring the limitations of analyses that rely on citation quantity alone and the potential for distortion of evidence.

\end{abstract}

\section{Introduction}

\begin{figure}[htbp]
    \centering
    \includegraphics[width=0.999\columnwidth]{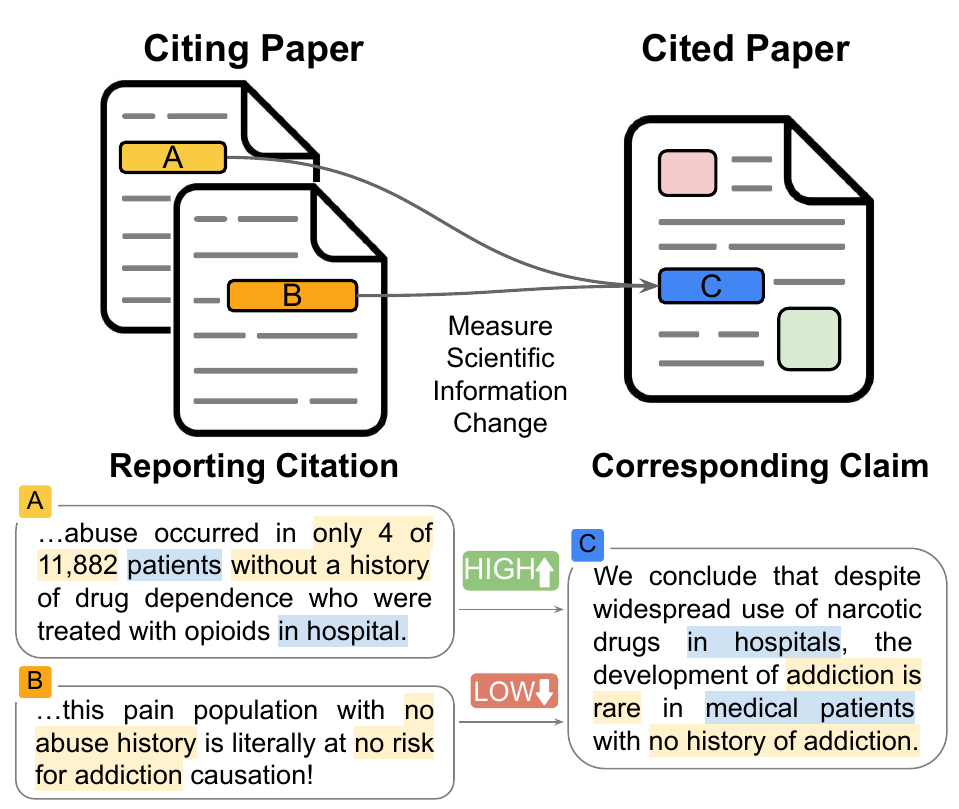}
    \caption{
    Citations do not always align with the original claims of cited works.
    Evidence on opioid addiction, originally restricted to a specific hospital setting, is often cited without critical contextual information. \protect\footnotemark 
     ~We develop a computational pipeline to assess citation \theword between citing and cited papers at scale.
    }
    \label{fig:figure1}
 \end{figure}

Citations are an essential part of scholarly communication. Given the assumption that authors cite prior work to acknowledge their intellectual debts to it \cite{kaplan1965norms,harter1992psychological}, citations are widely used to measure scientific contributions and trace the flows of knowledge. In practice, analysts and scholars tend to rely on raw citation counts \cite{waltman2016review}, implicitly assuming that citations are similar in kind, or that any variation is minor.  

However, a large literature has shown that citations vary substantially \cite{teufel-etal-2006-automatic, ValenzuelaEscarcega2015IdentifyingMC,zhu2015measuring,pride2017incidental, jurgens-etal-2018-measuring}. 

\footnotetext{
Example reporting citation (A) is from \citet{kowal1999issue}, and (B) is from \citet{portenoy1986chronic}. The corresponding claim (C) is sourced from \citet{porter1980addiction}. The low-\theword citation case overstates conclusions by failing to accurately specify the study population.}
Citations may paraphrase, summarize, or even misrepresent original knowledge, thereby leading to varying degrees of \textit{fidelity} between the original and the citing claims.
This variability in citation fidelity is not merely a theoretical concern; it has tangible consequences for the transmission of scientific knowledge. 
For instance, a short letter published in 1980 documenting opioid use in an inpatient sample \cite{porter1980addiction} was subsequently cited uncritically as evidence supporting the safe use of opioid prescriptions \cite{deyo2015opioids}. 
This letter had been cited over 600 times by 2017, with fewer than 20\% of citations acknowledging the critical information of the original context \cite{leung20171980}. 
Such citations collectively contributed to a misleading narrative of low addiction risk, which played a role in the subsequent opioid epidemic \cite{west2021misinformation}.

Despite the importance of citation \theword, it has been challenging to study at scale. In this study, we introduce a computational pipeline that overcomes the challenges. Specifically, using a large set of full-text research papers, we identify reporting citations from citing papers and their corresponding claims from cited papers, resulting in \num million cited-citing sentence pairs. We then apply a sentence-level measure of scientific information change to evaluate the citation \theword of each pair.

Our analysis reveals several patterns in citation \theword: \theword is higher when authors cite papers that are more recent and intellectually close and more accessible; 
within teams, citation \theword is highest for medium-sized teams, negatively correlated with the seniority of first authors, and shows no significant correlation with the seniority of last authors.
Additionally, we identify a ``telephone effect'' of intermediary citations. Specifically, using a quasi-causal experimental setting, we find that the \theword of citations is lower when the authors cite the original paper and an intermediary one that cites the original as well. Moreover, citation \theword is positively correlated with the \theword of the intermediary paper's citation to the original paper. This telephone effect suggests that intermediary engagement with original sources can result in information loss.\footnotemark 
\footnotetext{The codes and models for this paper are available at \url{https://github.com/hongcchen/citation}.}

\section{Motivation and Related Works}
Citations are a widely used bibliometric tool for evaluating the performance of various actors in the scientific community. To better understand citation behavior, researchers have increasingly examined the content of citations, especially the linguistic and contextual features of citing text, not only to infer authors' intent and citations' function, but also to interrogate a core assumption: to what extent do citations reflect genuine scientific influence on the citing authors and their work?

\subsection{Citation classification}

The normative theory of citation \cite{kaplan1965norms, merton1979normative} posits that authors cite papers to acknowledge their intellectual debts to them; thus citations should indicate a direct semantic relationship or relevance between citing and cited works \cite{garfield1979citation}. However, citation practices are not always accurate or rigorous, with issues such as misquotation, miscitation, and misrepresentation frequently observed \cite{Harzing2002AreOR, Horbach2021MetaResearchHP, cobb2024problem}.

A growing body of research has examined the relationship between citation and cited papers, revealing significant variations in citation \theword \cite{ west2021misinformation,broadus1983investigation,greenberg2009citation, zhang2024detecting}. Studies have investigated cases of citations deviating from the original findings with low \theword and have revealed that between 1 in 5 and 1 in 10 citations are used to support claims incongruous with the results of the cited paper \cite{todd2010one}. 
For example, \citet{cobb2024problem} examined 3,347 citing claims in top psychology journals and found that approximately 19\% either omitted crucial nuances or entirely misrepresented the claims of previous research. \citet{greenberg2009citation} showed how hypotheses can be transformed into accepted facts through affirmative citation chains lacking direct supporting evidence in biomedical research.
These findings reveal the variability in citation \theword and low fidelity in citation can contribute to the noisiness of the knowledge transmission pipeline. 
However, these studies are often limited in scale and reliant on labor-intensive manual verification which requires substantial domain expertise, leaving gaps in our understanding of citation \theword across broader academic discourse.

Existing citation text studies have scaled citation classification by automating the identification of citation intent or function, primarily leveraging the immediate context surrounding the reference or the metadata of the citing paper
\cite{cohan-etal-2019-structural, lauscher-etal-2022-multicite,pham2003new,li-etal-2013-towards,pride2017incidental,ValenzuelaEscarcega2015IdentifyingMC,teufel-etal-2006-automatic,baig-etal-2021-multitask}. For example, \citet{jurgens-etal-2018-measuring} classified the broad thematic functions of citations into six categories including 
\textsc{background}, \textsc{motivation}, \textsc{uses}, \textsc{extension}, \textsc{comparision or contrast} and \textsc{future}.
While these studies have expanded the scale of citation analysis, they still primarily focus on the citing paper, 
with limited efforts to use text of the \textit{cited} paper \cite{luu-etal-2021-explaining}, particularly for assessing citation fidelity.

Moreover, measuring citation fidelity extends beyond standard semantic textual similarity and requires assessing whether the citation faithfully conveys the same scientific information as the original text \cite{wright-etal-2022-modeling}. While there are potential alternative metrics from related areas such as fact-checking and semantic similarity \cite{thorne-etal-2018-fever, ganitkevitch-etal-2013-ppdb}, they are typically designed for different goals (e.g., truth verification, general paraphrase accuracy) and do not directly address fidelity in the context of scientific claims. This task of measuring fidelity requires capturing more nuanced aspects of similarity, such as abstraction level, uncertainty, and contextual appropriateness \cite{august-etal-2020-writing}.

In our work, we address this empirical gap in measuring citation fidelity by introducing a computational pipeline to examine this problem at scale. We analyze the relationship between citation pairs consisting of reporting citations and corresponding claims and use a measure of scientific information change to assess their \theword systematically.

\subsection{Author's engagement with literature}

A common criticism of using citations to indicate impact or trace knowledge flow is that authors may cite without directly engaging with the original source or being influenced by its content.
There is widespread allegation that scientists may not fully read or only superficially review the works they reference \cite{simkin2002read, jergas2015quotation}. \citet{kaplan1965norms} famously raised concerns about this practice, asking:
\textit{
“How often are the works of others cited without having been read carefully? 
How often are citations simply lifted from the bibliography in someone else’s work without either reading or giving credit to the man who did the original search of the literature?”
}

Existing research has primarily examined bibliographic errors as indicators of superficial engagement with the literature, which can contribute to questionable research practices.
For example, \citet{broadus1983investigation} found that 23\% of citing papers reproduced the same bibliographic error when using an incorrect title from a well-known sociology work, suggesting that many authors copy references without verifying the original sources. \citet{simkin2005stochastic} modeled the spread of misprints in bibliographies in physics and estimated that 70–90\% of citations are copied from other papers’ reference lists.

However, less attention has been paid to how authors’ engagement is reflected in subtle textual signals---specifically, how they report original findings in their citations. An author’s level of engagement may be revealed in how they reinterpret or modify original scientific information, whereas limited understanding or familiarity can result in lower alignment or fidelity.
Studies suggest that such low-fidelity citations may contribute to broader issues in knowledge transmission, such as the perpetuation of dominant discourses \cite{mizruchi1999social} or even the propagation of misinformation in science \cite{west2021misinformation}.

In this study, we conceptualize the \theword between citing and cited sentences as one reflection of authors' engagement with literature, defined as the depth of interaction an author has with the referenced work.
Prior studies have demonstrated that various factors can significantly influence how authors engage with and interpret prior research. 
For instance, temporal closeness has been shown to impact the depth of interaction, with recent and within-discipline works often fostering stronger connections \cite{medo2011temporal,wahle-etal-2025-citation}. Access barriers can also influence the depth of engagement with prior research \cite{evans2009open}.
At the same time, team dynamics, including team size and authorship roles, can affect how thoroughly prior studies are engaged and conveyed in writing, reflecting different levels of responsibility and expertise within collaborative research efforts \cite{lariviere2016contributorship, brand2015beyond}.

Building on these insights, we contribute to this question by hypothesizing and testing that \theword can be shaped by several key factors indicative of the level of engagement of authors, including temporal and disciplinary closeness, accessibility, and reliance on intermediary sources. These dimensions influence the depth of authors’ engagement with original literature, thereby affecting the fidelity of citations.

\section{Citation Sentence Pairs}

Examining citation \theword requires sentence-level analysis, so we leverage bibliometric datasets that provide both citation links and full-text content and extract citation sentence pairs---where the citing paper’s reporting sentence references a specific claim in the cited paper.

\subsection{Dataset}

In this study, we employ one of the most comprehensive academic datasets available: The Semantic Scholar Open Research Corpus \citep[S2ORC; ][]{lo-etal-2020-s2orc}. S2ORC is a general-purpose corpus containing metadata for 136 million scholarly articles spanning different fields. 
This dataset also provides access to the full text of articles, thus allowing us to analyze how authors make citations and identify which parts of the original text are being referenced on a large scale. After filtering out non-English articles and publications without full text, we obtained a dataset of approximately 42 million papers.
\subsection{Reporting sentences from citing papers}

Not all citations are intended to report the claims of prior literature. In this study, we focus on a specific category of citing sentences, referred to as reporting citations---sentences explicitly conveying the results or conclusions of a cited paper. To isolate these sentences, we apply two key filtering criteria: single-source citations and background citations.

\paragraph{Identifying single-source citations} We first locate the in-text citation by identifying the sentence where the reference appears, setting the citation context window to be a single sentence. \citet{li-etal-2022-corwa} find that many citations tend to correspond to a very localized piece of the cited document typically ranging from roughly a short clause to spans of one or two sentences. \citet{lauscher-etal-2022-multicite} also report that while references to prior literature can sometimes span multiple sentences, 83\% of citation contexts occur within a single sentence. Given that most citations fall within this single-sentence window, we restrict our analysis to single-sentence citations for consistency and precision.

As citations occur at varying levels of aggregation and granularity \cite{cronin1994tiered}, especially when multiple sources are cited together, we focus on citations that explicitly report claims from a single source. To further refine this selection, we apply a rule-based filter, considering only parenthetical citations enclosed in [] or () at the sentence’s end. This structured approach helps isolate the core reporting function of citations and minimizes confounding factors, enhancing the precision of our analysis. See Appendix Table~\ref{tab:reporting_citation} for valid examples of citing sentences.

\paragraph{Identifying background citations}
Citations can serve multiple functions beyond reporting claims from the original paper. Previous research has developed classification schemes for citations based on their roles within an argument. Among these, \textsc{background} citations are widely recognized as a category that provides essential context and relevant information for the domain under discussion \cite{teufel-etal-2006-automatic,cohan-etal-2019-structural,lauscher-etal-2022-multicite}. These citations often set the stage for a study by informing readers about existing knowledge, highlighting gaps, and establishing the context for the study’s purpose \cite{jurgens-etal-2018-measuring}.

Leveraging their framework, we focus on \textsc{background} citations as a means to further identify reporting sentences.
We construct a more comprehensive dataset by integrating annotated data from two prominent sources for citation function classification: SciCite \cite{cohan-etal-2019-structural} and MultiCite \cite{lauscher-etal-2022-multicite}, both of which annotate \textsc{background} as a citation function. The combined dataset includes 27,052 instances, with 11,635 (43\%) labeled as \textsc{background}. We fine-tune a SciBERT model \cite{beltagy-etal-2019-scibert} based on the integrated dataset for a binary classification task to identify whether a sentence serves as a \textsc{background} citation. The model achieves an F1 score of 0.81 using an 80-20 train-test split, which shows its effectiveness in accurately detecting \textsc{background} citations.

\subsection{Sentences of claims from cited papers}

To match reporting citations with the corresponding claims in cited papers, we follow the definition that background citations typically refer to relevant findings or essential context from the cited work. Therefore, we focus on sentences of claims, which are operationalized as those labeled as ``results'' or ``conclusions'' based on the discourse categories defined by \citet{dernoncourt-lee-2017-pubmed}. We apply a sentence classifier to identify result or conclusion sentences within the full text of the cited publication. This classifier is fine-tuned from a RoBERTa model \cite{liu2019robertarobustlyoptimizedbert} and is trained to categorize sentences into five categories: methods, background, objective, results, and conclusions. The training dataset consists of 200,000 paper abstracts from PubMed, which were self-labeled with these categories \cite{canese2013pubmed}. The classifier achieved an F1 score of 0.92 on a held-out 10\% sample \citep[details in ][]{wright-etal-2022-modeling}. Using this approach, we filter approximately 30\% of sentences of results or conclusions from the full text as claims to be matched with citing sentences.

\begin{table*}[h]  %
    \centering
    \renewcommand{\arraystretch}{1.5}  %
    \begin{tabular}{p{5.8cm}p{7.25cm} p{1.25cm}}  
        \textbf{Citing Sentence} & \textbf{Matched Sentence of Claim (Cited Paper)} & 
        \textbf{Fidelity} 
        \\ 
        \hline
        A large survey study showed that regular antenatal exposure to music and talking to the baby might prevent traits of ASD \textbf{[94]}.   & Our finding that exposing a fetus to music and maternal speech during pregnancy is associated with a broad reduction in autistic-like behaviors is congruent with previous findings of music training for children with ASD.   & 4.147   \\ 
        \hline
        One limitation in rsfMRI is intra-individual variability, which may be due to a variety of causes, such as time-of-day, diet, blood pressure, or cognitive load 
        \textbf{(Specht, 2019)}
        . & However, one major disadvantage of rs-fMRI is that rs-fMRI studies still vary in their acquisition methods and whether they are conducted on a 1.5T, 3T, or 7T MR.   & 2.817   \\ 
    \end{tabular}
    \caption{Examples of citing and matched sentences with \theword scores. The first pair from \citet{applewhite2022systematic} and \citet{ruan2018antenatal} shows high fidelity as both sentences discuss prenatal exposure to music and ASD, despite wording differences. The second pair from \citet{canario2021review} and \citet{specht2020current} shows lower \theword due to differing focuses on rs-fMRI limitations. 
    The second pair also reflects an interesting bibliographic inconsistency: the source is cited as 2019 though the paper was published in 2020.
    }
    \label{tab:table1}
\end{table*}

\section{Measuring Citation \textbf{Fidelity}}

Assessing citation \theword requires measuring how faithfully a citation reflects the original findings of the cited work. This involves first identifying the corresponding claims in the cited text, and then analyzing how the citation relates to that claim. Both are challenging due to the nuanced nature of scientific information and the absence of explicit sentence-level referential links in citations. To address this, we employ a sentence-level measure specifically designed for scientific findings, along with necessary approximations.

\subsection{Measuring sentence-level scientific information change}

Assessing \theword between citing and cited claims requires more than standard semantic similarity measures. It involves capturing nuanced aspects often inherent in scientific claims to determine whether the citation faithfully conveys the original information \cite{luu-etal-2021-explaining, wright-etal-2022-modeling, august-etal-2020-writing}.

For this purpose, we employ a sentence-level scientific information change measure developed by \citet{wright-etal-2022-modeling}. This measure quantifies the extent to which two semantically similar sentences describe the same scientific claims and capture topic alignment as well as both nuance and accuracy in how claims are reported. It is specifically designed for evaluating scientific texts and is well-aligned with our task of citation fidelity, with minimal domain or task shift.

The metric assigns a similarity score from 1 to 5, where 1 indicates that the two sentences describe completely different claims, and 5 indicates that the sentences express identical claims (see Appendix Table~\ref{tab:annotation}). The model is fine-tuned from MPNet \cite{10.5555/3495724.3497138} on a human-annotated dataset (crowdsourced and expert-reviewed) and trained to minimize the cross-entropy over sentence embedding dot products. It achieves a mean square error of 0.489 and a Pearson correlation of 76.48 in predicting the scientific information change \citep[details in ][]{wright-etal-2022-modeling}.

\paragraph{Upper-bound approach} 

Pairing the reporting citation and the corresponding claim is challenging because authors typically cite the entire publication rather than referring to specific sentences, making it difficult to determine which exact claim the citing sentence refers to. The absence of explicit sentence-level referential links complicates the task of assessing citation \theword.

To address this, we use an upper-bound approach, where the highest \theword score between a citing sentence and any sentence in the cited work serves as a proxy for \theword. This method identifies the best-matched corresponding claim in the original paper.
While the highest-scoring sentence may not always be the exact one the author intended to reference, this method approximates the best-case scenario of \theword and ensures that the most relevant claim is captured, providing a structured way to evaluate how much original information is represented in citations.

\begin{figure*}[t]
    \centering
    \includegraphics[width=0.99\textwidth]{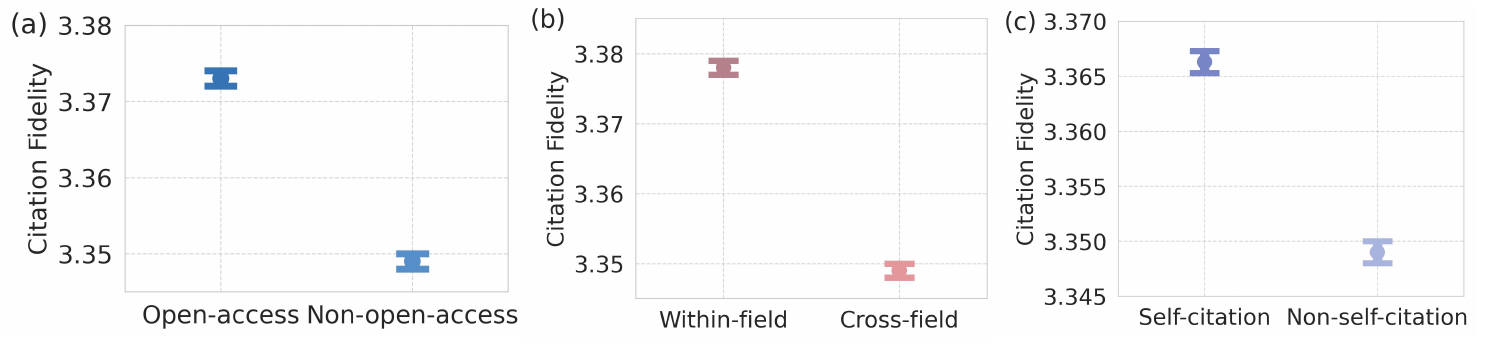}
    \caption{
    (a) Citation \theword is higher when citing within-field studies compared to cross-field studies.
    (b) Citation \theword is higher for self-citations compared to non-self-citations.
    (c) Citation \theword is higher when citing open-access works compared to non-open-access works.
    }
    \label{fig:2}
\end{figure*}

\subsection{Interpreting citation \theword}
Two examples of citation sentence pairs from our dataset are presented in Table~\ref{tab:table1}, illustrating how citation \theword quantifies the extent to which citing sentences accurately and comprehensively reflect the claims of cited papers. High-\theword citations demonstrate strong conceptual alignment, while low-\theword citations often diverge in the amount of information conveyed as well as in meaning, emphasis, or interpretation.

By extracting reporting citations and corresponding claims in cited sources, we generate \num million citation pairs. 
A random sample of fidelity score shows that the majority cluster around 3.5, following a roughly normal distribution ranging from 1.0 to 5.0 (See Appendix Figure~\ref{fig:distribution}). This suggests that most citations exhibit a reasonable level of \theword, though the presence of fewer extreme high and low values shows the variability in how accurately authors represent cited claims. 
Moderate \theword scores could reflect a balance between accurate representation and paraphrasing to fit the citing paper’s context, whereas lower scores might signal information loss or potential misrepresentation.

\section{Higher Fidelity is Associated with Closer Proximity}

How authors report claims when citing can vary based on factors such as disciplinary conventions, contextual relevance, and individual interpretation of original work. 
We are interested in how various factors---such as accessibility, and temporal and intellectual proximity---affect \theword. These factors, which can indicate the level of engagement an author has with the referenced work, may influence the alignment of information between citing and cited claims, as reflected in our fidelity measure.

\paragraph{Experimental setup} 
We employ a standard regression analysis framework, with the model specified below. Citation \theword is the dependent variable (\textit{F}). Key factors such as accessibility, proximity, and team dynamics are included along with other relevant attributes (See Appendix Table~\ref{tab:regression_variable} for more details on all variables). 
\begin{align*}
    \text{F} &= \beta_0 + \beta_1 \text{Field.of.Study} + \beta_2 \text{Publication.Year} \\
    &+ \beta_3 \text{Publication.Type} + \beta_4 \text{Open.Access} \\
    &+ \beta_5 \text{Context.Length} + \beta_6 \text{Reference.Frequency} \\
    &+ \beta_7 \text{Publication.Interval} + \beta_8 \text{Paper.Citation} \\
    &+ \beta_9 \text{Author.Seniority} + \beta_{10} \text{Team.Size}\\
    &+ \beta_{11} \text{Self.Citation}+ \beta_{12} \text{Within.Field}
\end{align*}

\begin{figure}[t]
    \centering
    \includegraphics[width=1.00\columnwidth]{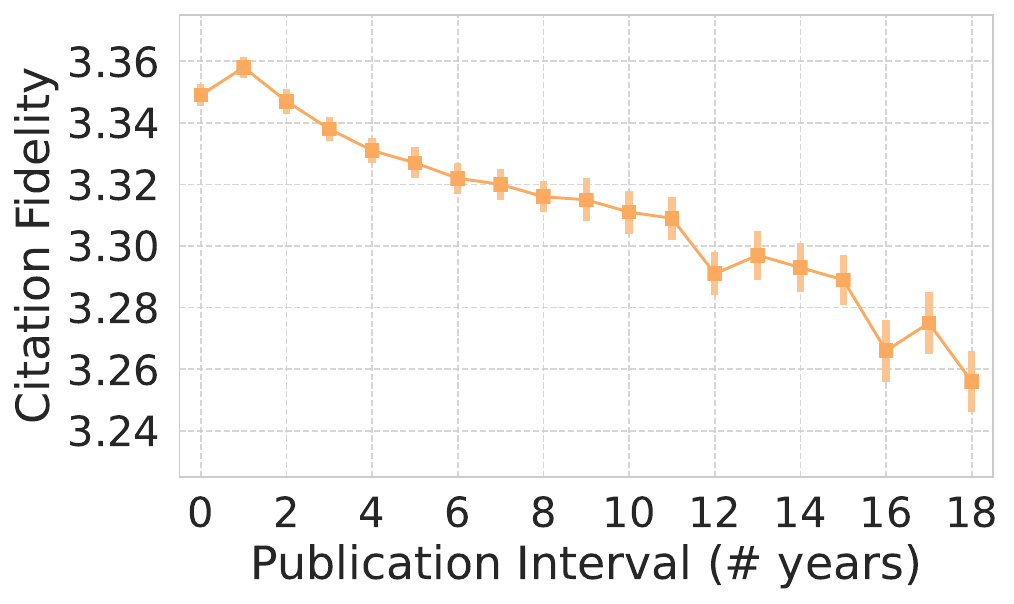}
    \caption{Citation \theword is higher when the publication gap between the citing and cited works is shorter. }
    \label{fig:3}
\end{figure}

\begin{figure}[t]
    \centering
    \includegraphics[width=0.99\columnwidth]{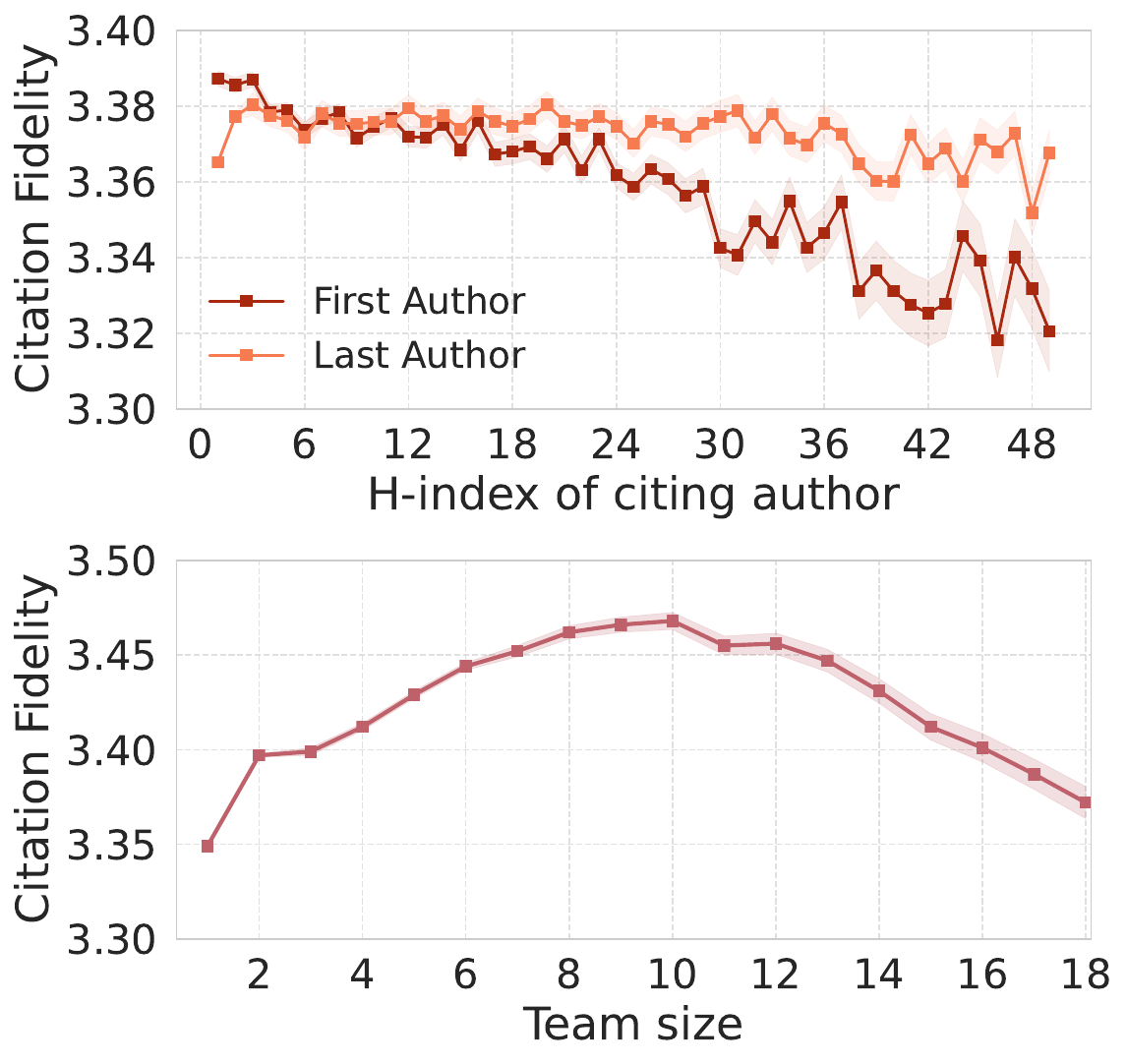}
    \caption{Citation \theword is negatively correlated with the seniority of first authors and less correlated with the seniority of last authors. Citation \theword is also higher for teams of medium size. }
    \label{fig:4}
\end{figure}

\paragraph{Results} Figures~\ref{fig:2}, \ref{fig:3}, and \ref{fig:4}
present coefficient estimates from regression analyses, controlling for potential confounders. The regression coefficients are mapped back to the original 0–5 fidelity scale by adding back the intercept, in order to visualize the results in a more interpretable way. Error bars represent standard errors (See Figure~\ref{fig:regressoin_result} for more regression results).
\footnote{Key variables are defined as follows: Team dynamics is proxied using \textit{Team.Size} and \textit{Author.Seniority}, where \textit{Team.Size} represents the number of authors in the team, and \textit{Author.Seniority} represents the experience of the authors using their H-index. Accessibility is proxied by \textit{Open.Access}, a binary variable indicating whether the cited paper is publicly accessible. Temporal and disciplinary proximity are proxied using \textit{Publication.Interval}, \textit{Self.Citation}, and \textit{With.Field}, where \textit{Publication.Interval} represents the number of years between the cited and citing papers, \textit{Self.Citation} indicates whether the two papers share at least one author, and \textit{With.Field} indicates whether they belong to the same discipline.}

First, our findings reveal that proximity---whether in terms of accessibility, disciplinary context, temporal closeness, or author familiarity---plays a critical role in shaping citation \theword. One strong piece of evidence is that self-citations, which is arguably a straightforward indicator of engagement as authors are naturally more familiar with their own work, exhibit higher \theword scores than non-self-citations. 
This supports the validity of our \theword measure, as it reflects the expected pattern: authors who engage more deeply with cited work tend to cite with higher \theword.

Similarly, open-access publications exhibit higher \theword compared to non-open-access publications, likely because unrestricted access to the original sources allows more direct engagement and accurate representation of cited work \cite{nagaraj2020improving}. Without access to the full content of the cited work, authors may only partially examine the original material, relying on overviews or others' interpretations. Citation \theword is higher for within-field citations compared to cross-field citations. This is likely because authors are more familiar with the methodologies, terminologies, or conceptual frameworks of their own disciplines, and are able to represent without much information change. In contrast, cross-field citations, while essential for interdisciplinary knowledge transfer \cite{thilakaratne-etal-2018-automatic}, may introduce additional information loss due to differences in their research norms, jargon, or methodologies. 

More recent publications tend to have higher \theword scores, likely because they are more contextually relevant and aligned with current scientific discourse. Authors may be more inclined to engage directly with recent works, while older publications are often cited less thoroughly \cite{wahle-etal-2025-citation}. It is often suggested that for canonical works, authors may rely on secondary interpretations rather than consulting the original source, which can lead to inaccuracies or information loss \cite{west2021misinformation}. We explore this dynamic further in Section~\ref{sec:telephone}.

Second, our findings suggest that the roles and collaborative dynamics within teams can shape how prior work is represented. Interestingly, citation \theword decreases as the seniority of first authors increases, while showing no significant relationship with the seniority of last authors---the traditional role for team leaders or principal investigators. This distinction may reflect the different roles and responsibilities within team collaboration. Last authorship often involves advisory or supervisory duties, where responsibility for writing detailed citations is typically delegated to other team members, usually the first author. \cite{brand2015beyond, sauermann2017authorship}.

The decline in citation \theword with increasing first-author seniority may indicate a shift in how senior researchers approach citations. More experienced authors might rely more on their established heuristics and prior knowledge, or adopt a more top-down writing approach, resulting in less engagement with the cited materials.

Citation \theword is highest for medium-sized teams, while both small and large teams exhibit relatively lower fidelity. This pattern likely reflects differences in team dynamics. Small teams may lack the diversity of expertise or necessary capacity needed for critical engagement with cited works, while in large teams, distributed responsibilities or coordination costs can reduce individual accountability for accurate representation \cite{lariviere2016contributorship, jones2021rise}. Medium-sized teams, however, may achieve a balance in between.

\begin{figure*}[t]
    \centering
    \includegraphics[width=1.0\textwidth]{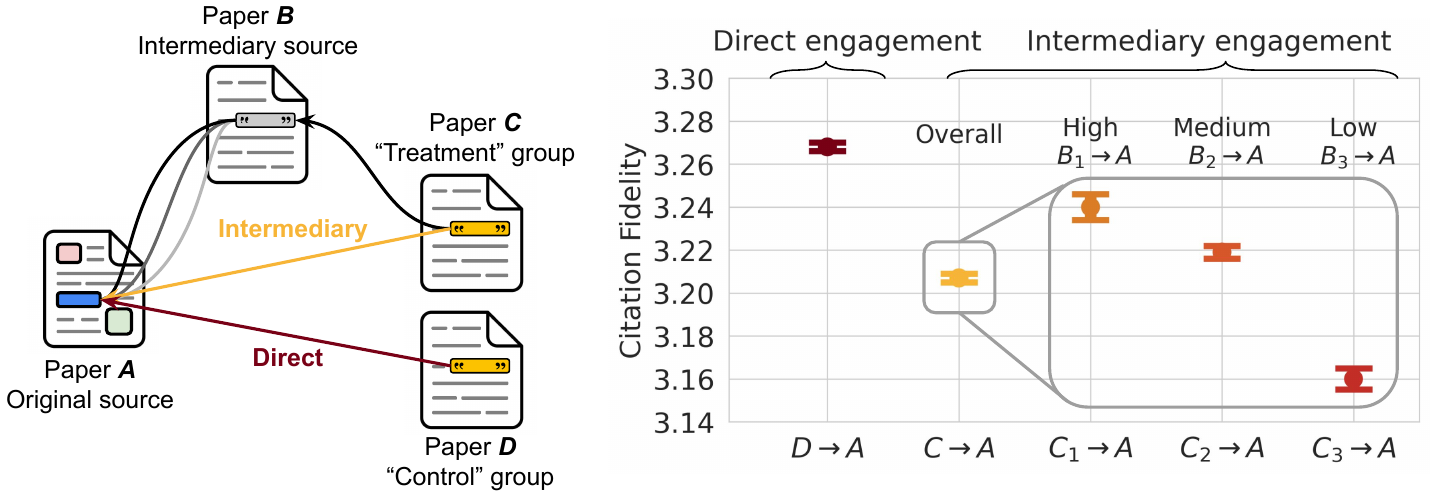}
    \caption{The left panel illustrates the citation relationship, where paper \textit{A} (original) is cited by paper \textit{C} and paper \textit{D}, with some authors also citing paper \textit{B} (treatment group). The right panel compares citation \theword to paper \textit{A} between the treatment group (authors citing both paper \textit{A} and paper \textit{B}) and the control group (authors citing only paper \textit{A}). Citation \theword in the treatment group is 0.06 lower than in the control group, suggesting that \theworda engagement with paper \textit{B} reduces \theword to paper \textit{A}. Error bars represent standard errors.}
    \label{fig:5-1}
\end{figure*}

\section{``Telephone Effect'' in Citation}
\label{sec:telephone}
A key concern in citation practices is that authors may not always engage directly with the sources they cite \cite{kaplan1965norms}. Instead, they often rely on intermediary sources---papers that have already interpreted and cited the original work. This intermediary engagement may introduce misinterpretations or distortions, which may propagate across the literature through citation chains \cite{simkin2002read, west2021misinformation}. Much like the ``telephone game,'' where a message is whispered from person to person and gradually distorted, scientific claims can evolve or degrade, as they are repeatedly paraphrased, summarized, or reinterpreted in subsequent research. We employ a quasi-causal experiment setting to examine how intermediary engagement with prior work impacts knowledge transmission, as reflected in their citation \theword.

We hypothesize that exposure to others' interpretations of an original claim influences how accurately a paper reports that claim. Specifically, citing an intermediary source increases the likelihood of information change from the original claim in the source paper, compared to when the paper directly engages with the source. This leads to two key modes of engagement: direct engagement (Paper \textit{C} cites Paper \textit{A} directly) and intermediary engagement (Paper \textit{C} cites Paper \textit{A} and an intermediary source Paper \textit{B}) that also cites Paper \textit{A}. See Figure~\ref{fig:5-1} for illustration.

To formalize this mechanism, we define the following hypotheses:
\begin{enumerate}
    \item \textbf{H1: Citation \theword decreases when authors rely on an intermediary source.} Specifically, one citation’s \theword in citing paper \textit{A} will be lower if they also cite paper \textit{B}, an intermediary source that also cites paper \textit{A}.
    \item \textbf{H2: The \theword of the intermediary source affects the \theword of subsequent citations.} If paper \textit{B} exhibits low \theword in citing paper \textit{A}, then a paper citing both A and \textit{B} will have lower citation \theword to \textit{A}.
\end{enumerate}

\paragraph{Experimental setup} To test these hypotheses, we conduct a quasi-experimental study, identifying approximately 50K paper pairs (\textit{C} and \textit{D}) with the following setup: both papers cite the same claim from same original paper \textit{A}, but paper \textit{D} cites \textit{A} directly (control group), while paper \textit{C} cites both \textit{A} and an intermediary paper \textit{B}, which itself cites \textit{A} (treatment group). To control for potential confounding factors, papers \textit{C} and \textit{D} are exactly matched by publication year and field, for comparability across both groups.

We use citations to an intermediary paper as a proxy for engagement with that paper. While this does not guarantee direct influence or even that the intermediary paper was read, it is associated with a higher chance of engagement. Consequently, citing an intermediary paper increases the likelihood of being influenced by its interpretation of the original claims compared to papers that do not cite any intermediary source.

\paragraph{Results} In the treatment group (citing both paper \textit{A} and paper \textit{B}), citation \theword to paper \textit{A} is 0.06 lower than in the control group (citing only paper \textit{A}). This finding supports \textbf{H1}, indicating that engagement with intermediary sources may contribute to low citation \theword. 

We further divide the treatment group based on the fidelity of the intermediary citation. Specifically, the fidelity of paper \textit{B}’s citation to paper \textit{A} is classified into three levels: high (score above 4), medium (scores of 3–4), and low (score below 3). We find the \theword of paper \textit{C}’s citation to paper \textit{A} positively correlates with the \theword of paper \textit{B}’s citation to paper \textit{A}, corroborating \textbf{H2}. Low citation \theword in the intermediary source can lead to further information loss in subsequent citations. 

These results show the inherent risks associated with relying on intermediary sources in scholarly communication. While intermediaries often serve as practical and time-efficient tools for navigating the large body of scientific literature, they can also become vectors for information loss or even misrepresentation \cite{sharkey2023expert}. This is particularly concerning when intermediaries fail to accurately reflect the original claims, as their interpretations may cascade through subsequent citations, compounding distortions over time \cite{west2021misinformation}. 
Such reliance not only affects the \theword of knowledge transmission but also raises broader questions about the robustness of scholarly practices. 

Our quasi-experimental results reflect the average effect of having any intermediary source on citation fidelity. However, intermediary effects are likely more nuanced. The number of intermediaries involved, as well as their characteristics, such as recency, prestige, and citation count, may differentially influence how interpretations are transmitted. Future work could explore these dimensions to better understand the role intermediary sources play and investigate the specific mechanisms through which these low fidelity occur, such as affirmative citation or selective reporting to understand the source of low \theword.

\section{Conclusion}

In this study, we introduced a computational pipeline to analyze citation \theword on a large scale and developed an automated measure to evaluate how authors report prior findings. By applying this method to \num million citation sentence pairs, we uncovered several key insights into the dynamics of scholarly communication and showed citation fidelity is higher with closer proximity. Through a quasi-causal experiment, we further establish that the ``telephone effect'' contributes to information loss: when citing papers exhibit low \theword to the original claim, future papers that cite both the intermediary and the original show even lower \theword to the original.

This study addresses two critical research gaps: the lack of large-scale measurements to assess citation \theword and the lack of empirical evidence on how authors engage with prior literature. By operationalizing citation \theword and exploring its contextual drivers, we reveal systematic differences in citation \theword and provide a more nuanced understanding of how information is preserved or lost during the citation process. These findings not only provide evidence on the mechanisms of knowledge transmission but also reveal the limitations of traditional citation analyses which often overlook dimensions of \theword and authors' engagement.

Finally, this study represents an initial step towards enriching citation analysis with text-level nuance. It could help advance our understanding of citation quality---as a potential source of distorted or misrepresented information, as a behavioral trace of author engagement with prior work, and as a valuable complement to existing metrics for research evaluation.

\section{Limitations}
Citations are made for different purposes and with different functions, and authors employ a variety of ways in writing to paraphrase, summarize, or reinterpret prior work. This variety makes systematically measuring citation fidelity challenging. In our analysis, we apply a series of filters to both citing and cited sentences to form appropriate citation pair instances---for example, limiting citing sentences to those with a single source cited at the end, and limiting cited sentences to single-sentence claims labeled as results or conclusions. These filters help us isolate citation pairs where claims are expected to be reported faithfully. They prioritize precision but at the expense of recall and they may also introduce potential selection bias and overlook alternative citation forms that are also interesting to investigate.

In particular, we restrict the analysis to one-to-one, sentence-level matches between citing sentences and candidate claims in the cited paper, with the fidelity score proxied as the highest alignment of each sentence pair. However, in practice, authors may distribute their reporting text across multiple sentences or use only a brief clause. Likewise, cited claims may also be diffuse, spanning several segments of text in the cited work. As a result, this simplification could likely underestimate fidelity in more complex rhetorical structures. Future work could improve upon this by expanding the scope of citation context and claim spans.

Also, while our citation \theword scores capture alignment across several dimensions---such as topical alignment, generality, and factual representation of the original claim---they reflect an aggregate measure rather than disentangling these specific components. So these scores may oversimplify nuanced aspects of citation behavior and may be influenced by factors such as varying levels of abstraction, synthesis with other sources, or factual inaccuracies in reported claims. These factors may blur the boundary between acceptable paraphrasing and substantive misrepresentation. Therefore, while our fidelity measure represents a meaningful first step toward identifying potential sources of misrepresented scientific evidence, it does not correspond directly to the widely discussed concept of ``miscitation'' or ``reference error.'' Future research should explore these finer-grained aspects, not only to identify such cases but also to gain deeper insights into how and why misinformation or questionable research practices emerge and propagate through the citation process.

\section{Acknowledgments}
We thank members of Blablablab and DiscoveryLab, as well as participants in the Computational Social Science Working Group at the University of Michigan, for their insightful feedback and suggestions on this project.

\bibliography{custom}

\newpage

\appendix

\section{Appendix}
\label{sec:appendix}

\subsection*{Inference and Training Details}
\label{sec:inferencedetail}
All experiments were conducted on a single NVIDIA RTX A6000 GPU using HuggingFace Transformers 4.36.2 and PyTorch 2.1.2 in a CUDA 12.1 environment. 

We trained a model for the \textsc{background} citation classification task. For fine-tuning the SciBERT model on the combined dataset, we trained for 10 epochs with a learning rate of \( 10^{-5} \), 500 warmup steps, and a batch size of 16. The best model was selected based on the validation loss.

The fine-tuning process was computationally efficient, taking approximately 20 minutes to complete on the dataset of 27,052 annotated instances. For inference, the trained model processes citations at a rate of approximately 15K instances per minute, allowing for rapid annotation of large-scale citation data. This setup ensures both scalability and accuracy for identifying \textsc{background} citations in citation analysis tasks.

\subsection*{Examples of Reporting Citation}

Table~\ref{tab:reporting_citation} presents examples of citing sentences that meet our criteria for single-source reporting citations. These cases are characterized by citations appearing at the end of a sentence, enclosed in brackets or parentheses, and explicitly attributing a claim to one single source.

By filtering for sentences dedicated to reporting findings from a single source, we create a controlled and precise dataset that isolates the core reporting function of citations, improving the reliability of our analysis.

\begin{table}[h]  %
    \centering
    \renewcommand{\arraystretch}{1.5}  %
    \begin{tabular}{p{5.8cm}p{1.0cm}}  
        \textbf{Citing sentence} & \textbf{Valid?} 
        \\ 
        \hline
        \dots past work has shown that active contributors in r/science are largely already involved in scientific activity [17] \dots    & Yes  \\ 
        \hline
        \dots Existing studies in NLP to help automate the study have examined exaggeration [17], certainty [18], and fact checking [19], among others \dots  & No  \\
        \hline
        \dots We use GROBID [12], a more commonly used and actively developed tool \dots   & No  \\
        \hline
        \dots The finding was in accordance with former studies (Lee et al. 2020) \dots  & No  \\ 
    \end{tabular}
    \caption{We focus on a specific category of citing sentences that feature single-source background citations, with the reference placed at the end of the sentence. Valid examples of reporting citations are shown.}
    \label{tab:reporting_citation}
\end{table}  %

\subsection*{Measure of Scientific Information Change}

To assess the fidelity of citations, we employ a supervised model to evaluate the degree of scientific information change between two claims. This model is trained on annotated data that evaluate whether two scientific claims convey the same information. Table~\ref{tab:annotation} outlines the annotation framework used to quantify scientific information change. The metric assigns a score from 1 to 5, where higher scores indicate greater alignment between the claims. Annotators were instructed to assess similarity based on the conveyed scientific content rather than surface-level semantic similarity. For further details on the annotation process, see \citet{wright-etal-2022-modeling}.

\begin{table}[H]  %
    \centering
    \renewcommand{\arraystretch}{1.5}  %
    \begin{tabular}{p{5.8cm}p{0.95cm}}  
        \textbf{Annotation Instruction} & \textbf{Score} 
        \\ 
        \hline
        The information in the findings is completely the same.   & 5  \\ 
        \hline
        The information in the findings is mostly the same.  & 4  \\
        \hline
        The information in the findings is somewhat similar.  & 3  \\
        \hline
        The information in the findings is mostly different.  & 2  \\ 
        \hline
        The information in the findings is completely different.  & 1  \\ 
    \end{tabular}
    \caption{The metric assigns a similarity score from 1 to 5, where 1 indicates completely different scientific claims, and 5 indicates identical scientific claims.}
    \label{tab:annotation}
\end{table}  %

\subsection*{Citation Fidelity Distribution}
\begin{figure}[H]
    \centering
    \includegraphics[width=.99\columnwidth]{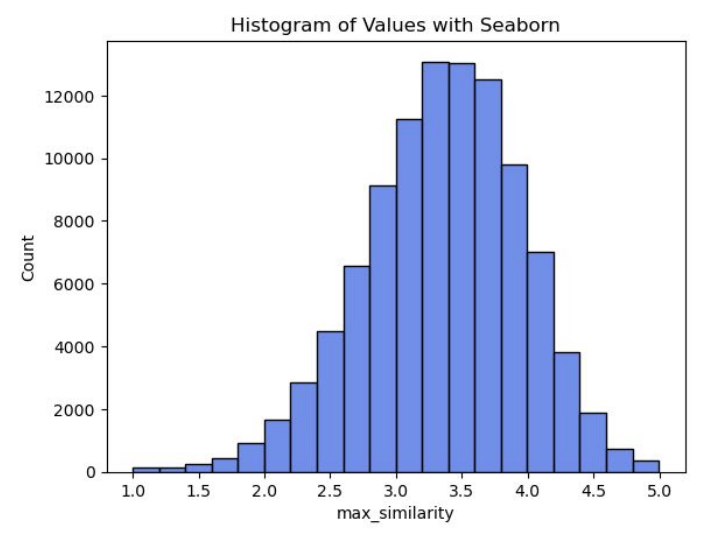}
    \caption{Distribution of fidelity scores across the dataset. A random 10k sample of fidelity scores reveals that the majority cluster around 3.5, following a roughly normal distribution ranging from 1.0 to 5.0.}
    \label{fig:distribution}
\end{figure}

Figure~\ref{fig:distribution} presents the distribution of citation fidelity across a random 10k sample from our dataset of sentence pairs. The histogram reveals a roughly normal distribution, with most scores clustering around 3.5. While a substantial portion of citations maintain a reasonable level of alignment with the original claims, the presence of both high- and low-fidelity cases shows variability in citation practices. 

\section{Regression Analysis Details and Results}

We use a regression framework to analyze factors influencing citation \theword, with \theword as the dependent variable. Descriptions of all variables, including key factors (such as \textit{Open.Access}, \textit{Within.Field}, etc.) and control variables (such as \textit{Field.of.Study}, \textit{Publication.Type}, \textit{Reference.Frequency}, etc.), are provided in Table~\ref{tab:regression_variable}. 

\begin{figure*}
    \centering
    \includegraphics[width=2\columnwidth]{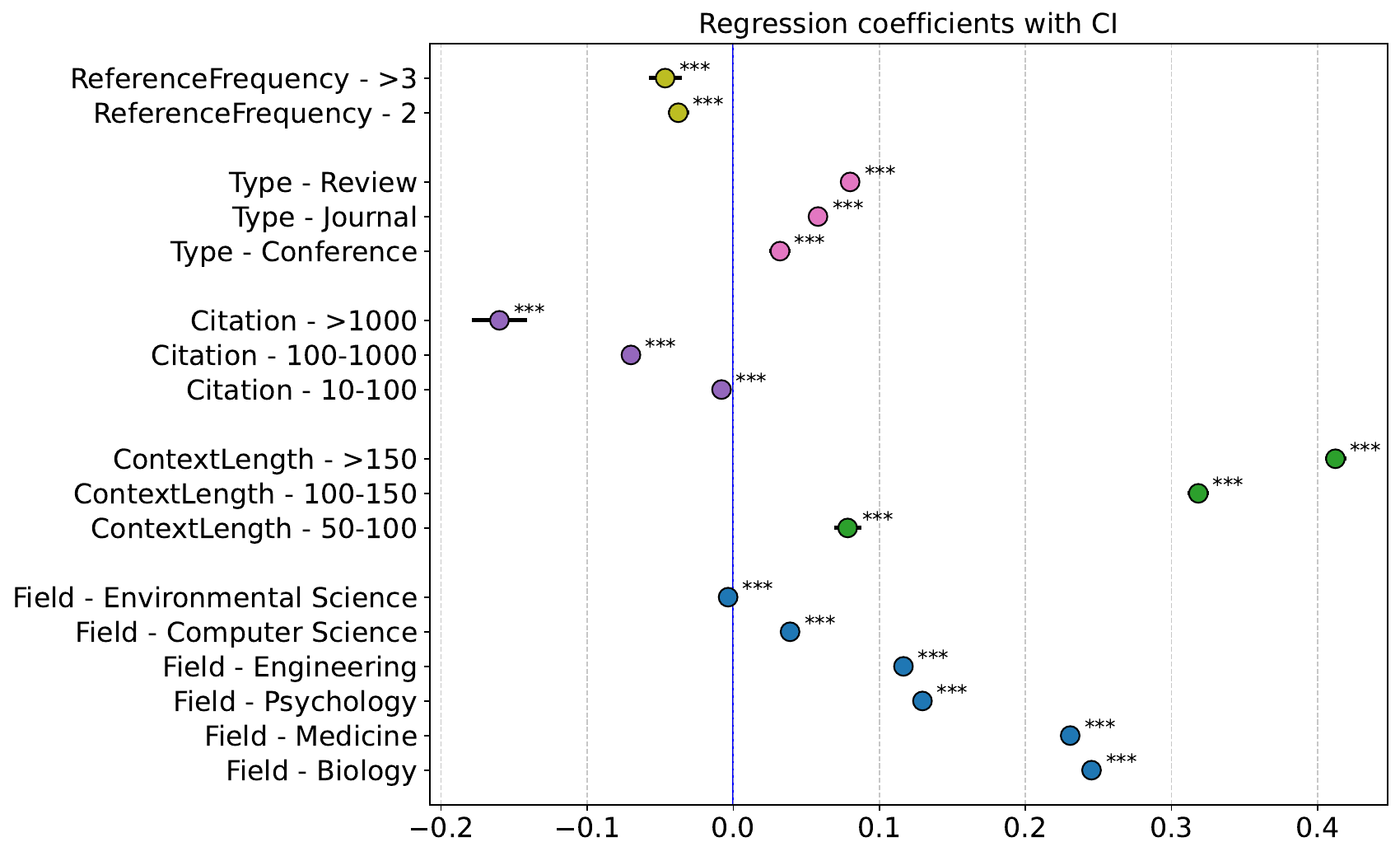}
    \caption{Regression coefficients with 95\% confidence intervals for predictors of citation \theword. Each point represents the estimated effect of a predictor (e.g., discipline,  publication type, context length) on citation \theword. Horizontal lines indicate the 95\% confidence intervals for each coefficient. The statistical significance level is indicated by asterisks (*** for p < 0.001).}
    \label{fig:regressoin_result}
\end{figure*}

Figure~\ref{fig:regressoin_result} presents the regression coefficients for additional factors (beyond those discussed in Section 5) affecting citation \theword, with 95\% confidence intervals. Categories within predictors (e.g., \textit{Field.of.Study} and \textit{Publication.Type}) are color-coded for clarity. The reference categories for categorical predictors are as follows: \textit{``Field: Physics,'' ``OpenAccess: False,'' ``Publication Type: Other,'' ``Publication Year: 2000,''  ``Within-Field: False,'' and ``Self-Cited: False.''} For continuous variables, they are grouped into bins for visualization purposes to better illustrate trends (in these cases, the bin corresponding to the smallest value serves as the reference category).

\begin{table*}[htp]
    \centering
    \renewcommand{\arraystretch}{1.5}  %
    \begin{tabular}{p{4cm}p{11cm}}  
        \hline
        \textbf{Variable} & \textbf{Description} \\ 
        \hline
        \textit{Field.of.Study} & The assigned research discipline of the citing paper, as provided by the S2ORC dataset. S2ORC determines this by leveraging Microsoft Academic Graph fields of study, which are automatically assigned based on metadata and citation relationships. These fields represent the primary discipline of the paper (e.g., computer science, biology).  \\  
        \hline
        \textit{Publication.Year} & The year in which the citing paper was published. This is treated as a categorical variable to capture temporal trends in citation behavior. \\ 
        \hline
        \textit{Publication.Type} & A categorical variable indicating the document type or genre of the citing paper, such as journal articles, conference papers, review, etc. \\  
        \hline
        \textit{Open.Access}  & A binary variable indicating whether the cited paper is publicly accessible. \\  
        \hline
        \textit{Context.Length}  & The number of characters in the citing sentence containing the reference. This captures the verbosity of the citation context. \\  
        \hline
        \textit{Reference.Frequency}  & The number of times the same cited paper appears within the citing paper. This quantifies the prominence of a reference within a document. \\  
        \hline
        \textit{Publication.Interval}  & The time difference (in years) between the publication dates of the cited and citing papers, which captures the recency of citations. \\  
        \hline
        \textit{Paper.Citation}  & The total number of citations that the cited paper has received at the time of dataset compilation. This is obtained from S2ORC’s citation metadata. \\  
        \hline
        \textit{Author.Seniority}  & The seniority of the authors of the citing paper, measured using the highest H-index among the listed authors. This is obtained from S2ORC’s author metadata. \\  
        \hline
        \textit{Team.Size}  & The number of authors listed on the citing paper. \\  
        \hline
        \textit{Self.Citation}  & A binary variable indicating whether the citing and cited papers share at least one author. \\  
        \hline
        \textit{Within.Field}  & A binary variable indicating whether the citing and cited papers belong to the same \textit{Field.of.Study}. \\  
        \hline
    \end{tabular}
    \caption{Descriptions of independent variables used in our analysis. These variables are either directly obtained from S2ORC’s metadata or derived based on attributes within the dataset. See \citet{lo-etal-2020-s2orc} for details on how S2ORC constructs metadata.}
    \label{tab:regression_variable}
\end{table*}

Besides the major findings related to \theword, disciplinary differences were also observed. Citation fidelity is higher in fields such as Biology and Medicine compared to Physics and Computer Science. This suggests that researchers in the life sciences may engage more closely with cited works. In contrast, fields like Computer Science, where rapid paradigm shifts and fast dissemination are common, exhibit lower levels of fidelity.

Publication type can also influence citation fidelity. Review articles demonstrate the highest fidelity, followed by journal articles, while conference papers tend to show lower fidelity. This pattern may reflect differences in citation reporting standards and conventions across publication venues. Review articles, by design, are expected to more accurately report existing claims with higher fidelity in their citations. And conference papers, which often emphasize novelty and prioritize brevity, may engage with prior work in a more flexible way with lower fidelity.

Citation fidelity decreases as the citation count of the cited paper increases. This trend suggests that heavily cited papers are often engaged with less deeply, possibly due to their canonical status or the reliance on a large number of secondary interpretations. Papers with high citation counts may be cited symbolically or to signal familiarity rather than as direct sources of content, leading to a low fidelity in their representation.

Citation fidelity also increases with longer citation contexts. However, the increase diminishes beyond approximately 100 words, indicating that longer contexts are associated with more thorough explanations of the cited work, but excessively long contexts may dilute focus or introduce irrelevant details. 
Similarly, frequent citation of a source within a paper is associated with lower fidelity, suggesting more extensive interpretation and paraphrasing of the cited content. When a source is referenced multiple times, authors may engage with different aspects of the cited work rather than reiterating the same claim, leading to greater variation in fidelity.

\begin{table*}[htbp] \centering
\begin{tabular}{@{\extracolsep{15pt}}lc}
\\[-1.8ex]\hline
\hline \\[-1.8ex]
\textbf{Variables} & \textbf{Coefficient} \\ 
\hline \\[-1.8ex]
 Intercept & 3.348$^{***}$  \\
 Field.of.Study &  \\
 \qquad Biology &  0.245$^{***}$\\
 \qquad Medicine &  0.231$^{***}$\\
 \qquad Psychology &  0.129$^{***}$\\
 \qquad Engineering &  0.116$^{***}$\\
 \qquad Computer Science &  0.039$^{***}$\\
 \qquad Environmental Science &  -0.003$^{***}$\\
 Publication.Type &  \\
 \qquad Review & 0.080$^{***}$  \\
 \qquad Journal & 0.058$^{***}$  \\
 \qquad Conference & 0.032$^{***}$  \\
 Open.Access & 0.024$^{***}$  \\
 Context.Length & 0.003$^{***}$  \\
 Reference.Frequency & -0.016$^{***}$  \\
 Publication.Interval & -0.007$^{***}$  \\
 Paper.Citation & -4.10\text{e}{-6} $^{***}$  \\
 Author.Seniority & -0.001$^{***}$  \\
 Self.Citation & 0.017$^{***}$  \\
 Team.Size & 6.86\text{e}-6  \\
 Within.Field & 0.029$^{***}$  \\
\hline \\[-1.8ex]
 Publication.Year & Yes \\
 Observations & 13587728 \\
 $R^2$ & 0.054 \\
 Adjusted $R^2$ & 0.053 \\
\hline
\hline \\[-1.8ex]
\textit{Note:} & \multicolumn{1}{r}{$^{*}$p$<$0.05; $^{**}$p$<$0.01; $^{***}$p$<$0.001} \\
\end{tabular}
\caption{Regression results examining factors associated with variation in citation \theword. The model estimates the relationship between \theword and a set of explanatory variables. Significance levels indicated by asterisks.}
\end{table*}

\end{document}